\def\year{2022}\relax
\documentclass[letterpaper]{article} % DO NOT CHANGE THIS
\usepackage{aaai22}  % DO NOT CHANGE THIS
\usepackage{times}  % DO NOT CHANGE THIS
\usepackage{helvet}  % DO NOT CHANGE THIS
\usepackage{courier}  % DO NOT CHANGE THIS
\usepackage[hyphens]{url}  % DO NOT CHANGE THIS
\usepackage{graphicx} % DO NOT CHANGE THIS
\usepackage{natbib}  % DO NOT CHANGE THIS AND DO NOT ADD ANY OPTIONS TO IT
\usepackage{caption} % DO NOT CHANGE THIS AND DO NOT ADD ANY OPTIONS TO IT
\DeclareCaptionStyle{ruled}{labelfont=normalfont,labelsep=colon,strut=off} % DO NOT CHANGE THIS
\usepackage{algorithm}
\usepackage{algorithmic}

\usepackage{enumitem}
\usepackage{makecell}
\usepackage{xspace}
\usepackage{amsmath,amssymb,amsfonts}
\usepackage{graphicx}
\usepackage{textcomp}
\usepackage{subfigure}
\usepackage{tcolorbox}
\usepackage{booktabs}
\usepackage{tabularx}
\usepackage{lipsum}
\usepackage{multirow}
\usepackage{color, xcolor}

\def\ie{\textit{i.e.}}

\newcommand{\fig}[1]{Fig.~#1}
\newcommand{\tab}[1]{Table~#1}
\newcommand{\eqn}[1]{Equation~#1}

\definecolor{red}{rgb}{0.8823529411764706,0.0745098039215686,0.3568627450980392} %{225,19,91}
\definecolor{green}{rgb}{0.1529411764705882,0.6862745098039216,0.403921568627451}   %{39,175,103}
\definecolor{blue}{rgb}{0.0470588235294118,0.4156862745098039,0.6156862745098039}
\definecolor{lightGreen}{rgb}{0.3764705882352941,0.7568627450980392,0.7294117647058824}
\definecolor{black}{rgb}{0,0,0}

\usepackage[switch]{lineno}
\usepackage{newfloat}
\usepackage{listings}
\floatstyle{ruled}
\newfloat{listing}{tb}{lst}{}
\floatname{listing}{Listing}
 \usepackage{mathrsfs}
\title{
Reliable Propagation-Correction Modulation for Video Object Segmentation
}
\author{
    Xiaohao Xu, \textsuperscript{\rm 1,2}\thanks{The work was done when Xiaohao Xu was an intern at MSRA.}
    Jinglu Wang, \textsuperscript{\rm 2}
    Xiao Li,\textsuperscript{\rm 2}
    Yan Lu \textsuperscript{\rm 2}
}

\affiliations{
    \textsuperscript{\rm 1} Huangzhong University of Science \& Technology\\
    \textsuperscript{\rm 2} Microsoft Research Asia\\
    xxh11102019@outlook.com, \{jinglwa, xili11, yanlu\}@microsoft.com
}

\usepackage{bibentry}
\begin{document}

\maketitle

\begin{abstract}
Error propagation is a general but crucial problem in online semi-supervised video object segmentation.
We aim to suppress error propagation through a correction mechanism with high reliability.
The key insight is to disentangle the correction from the conventional mask propagation process with reliable cues.
We introduce two modulators, \textit{propagation} and \textit{correction modulators}, to separately perform channel-wise re-calibration on the target frame embeddings according to local temporal correlations and reliable references respectively.
Specifically, we assemble the modulators with a cascaded \textit{propagation-correction} scheme. This avoids overriding the effects of the reliable correction modulator by the propagation modulator. 
Although the reference frame with the ground truth label provides reliable cues, it could be very different from the target frame and introduce uncertain or incomplete correlations. We augment the reference cues by supplementing reliable feature patches to a maintained pool, thus offering more comprehensive and expressive object representations to the modulators. In addition, a reliability filter is designed to retrieve reliable patches and pass them in subsequent frames.
Our model achieves state-of-the-art performance on YouTube-VOS18/19 and DAVIS17-Val/Test benchmarks.
Extensive experiments demonstrate that the correction mechanism provides considerable performance gain by fully utilizing reliable guidance. Code is available at:  \textit{https://github.com/JerryX1110/RPCMVOS}.
\end{abstract}

\section{Introduction}
\label{sec:introduction}
\begin{figure}[t]
\centering                                  
\includegraphics[width=0.45\textwidth]{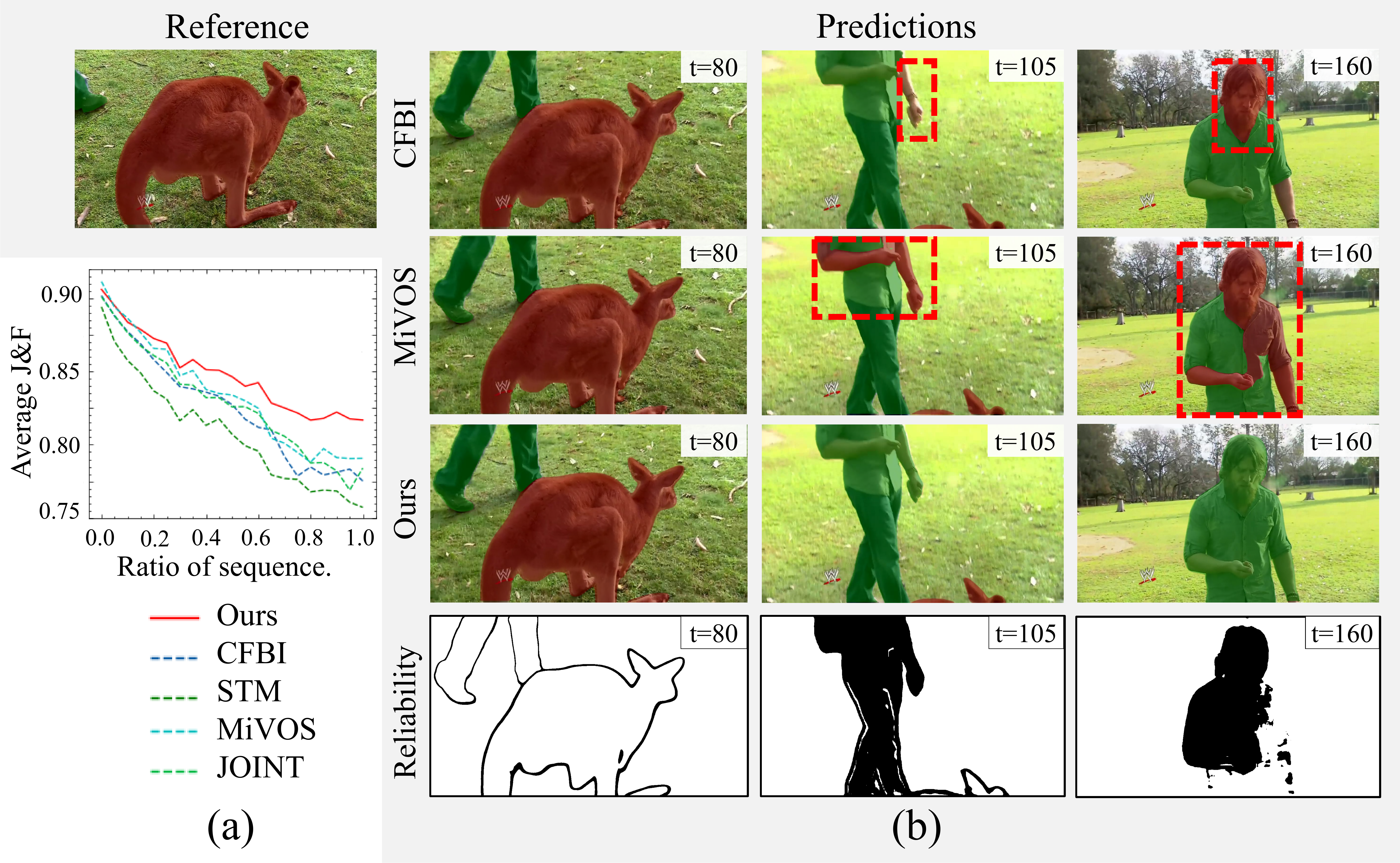}%,height=3.00cm

\caption{Our model can suppress error propagation in VOS with a reliable propagation-correction mechanism. (a) Average overall performance (J\&F) on YouTube-VOS 19 validation set over time. Ours has the least performance decay. (b) Considering the large appearance transition from the reference to targets, our model achieves much better results compared to CFBI \cite{yang2020collaborative} and MiVOS \cite{cheng2021mivos} (red rectangles bound error regions). The binary reliability maps indicate reliable (white) and uncertain (black) patches.}

\label{fig:teaser}                   
\end{figure}

Semi-supervised video object segmentation (VOS), also known as mask tracking, aims at segmenting target objects in a video sequence given the ground truth mask at the first (or reference) frame. Recently, sequence-to-sequence methods \cite{vaswani2017attention,duke2021sstvos} achieve impressive results but suffer from relatively high cost. Online methods \cite{perazzi2017learning,oh2018fast,wang2018semi,wang2019ranet}, taking only the current frame with image-wise references as input, are more practical for fast and streaming applications. We focus on online methods in this paper. 

The Semi-supervised VOS problem is usually formulated as a maximum a posterior (MAP) problem, conditioning on the target frame, preceding and reference frames and labels. 
Considering the probabilistic model of online VOS, the current label can be predicted from a frame-by-frame propagation path or a direct translation path from the reliable reference label. 
To exploit local temporal consistency, many methods \cite{oh2018fast,perazzi2017learning,NIPS2017_6c9882bb,voigtlaender2017online,cheng2017segflow,cheng2018fast,NIPS2017_6c9882bb} follow the propagation path to perform mask propagation, but errors may accumulate over time due to the inevitable prediction uncertainty in each iteration. The reference frame with a ground truth label provides reliable object cues, thus having the potential to suppress error propagation \cite{NEURIPS2020_0d5bd023}. Recent methods \cite{voigtlaender2019feelvos,yang2020collaborative} demonstrate that even naively manipulating references by feature concatenation and matching could improve the VOS performance. This encourages us to make full use of reliable reference cues to correct errors during the mask propagation.
However, the target frame may turn out to be very different from the reference when time goes by, losing explicit correspondences to the reference. For example, in \fig{\ref{fig:teaser}} (b), the reference only containing part of the object is not comprehensive to represent the whole object, \ie, the foot in reference. In this case, estimating correlations between the reference and target is uncertain and incomplete, which may lead to negative impacts on the VOS task.

Network modulation, which recalibrates feature embeddings with additional conditions, has achieved great success in VOS \cite{Yang2018osmn,yang2020collaborative}. The modulation operation is light-weight and can be performed per frame, which fulfils the streaming requirement.
The key to modulation is to construct expressive conditional weights and extract highly correlated embeddings. For the VOS task, modulation weights need to be representative for reference objects and embeddings also need to encode reliable correlations between the reference and the target.
In this paper, we propose a new end-to-end framework for VOS with reliable propagation-correction modulation, which can provide representative object proxies weights for modulation and consolidate target embeddings in a cascaded assembly of propagation and correction modulators. 

To perform correction with high reliability, we augment the translation path from the reference to a more comprehensive correction path.
Since object cues in the reference frame may be incomplete, we progressively supplement them with reliable information in each iteration. A reliable patch memory pool is maintained to store the historical reliable feature patches, which is further utilized in subsequent frames. The reliable patch pool is for two usages, \ie, augmenting the object proxy with comprehensive information to obtain expressive modulation weights and consolidating frame embedding with more reliable correlations.  
As for the network design, we introduce two types of modulator, \ie, propagation and correction modulator, which separately augment embedding according to the local temporal correlation cues in the propagation path and the reliable reference cues from the correction path. To avoid overriding the correction effect, the correction modulator is inserted after the propagation modulator.
We also propose a reliability filter to assess the prediction quality. From the example reliability map in \fig{\ref{fig:teaser}} (b), regions with large appearance change from reference are predicted as uncertain while other reliable regions can be passed to the following frames for correction.
Our experiments demonstrate that the assembly of propagation and correction modulators has a considerable impact on VOS performance. 
Our contributions are three-fold. 

\begin{itemize}
    \item 
    We propose a new reliable propagation-correction modulation model for VOS, which significantly suppresses error propagation (see precision decay curve in \fig{\ref{fig:teaser}} (a)). Our model achieves the state-of-the-art performance on both YouTube-VOS and DAVIS17 benchmarks.
    \item We disentangle the reliable correction from the conventional erroneous mask propagation process with separate memory modulators.
    \item We augment both the object proxy and target embedding with comprehensive reliable cues to reinforce the correction modulation.
\end{itemize}

\begin{figure*}[t]
	\centering
	\includegraphics[width=\textwidth]{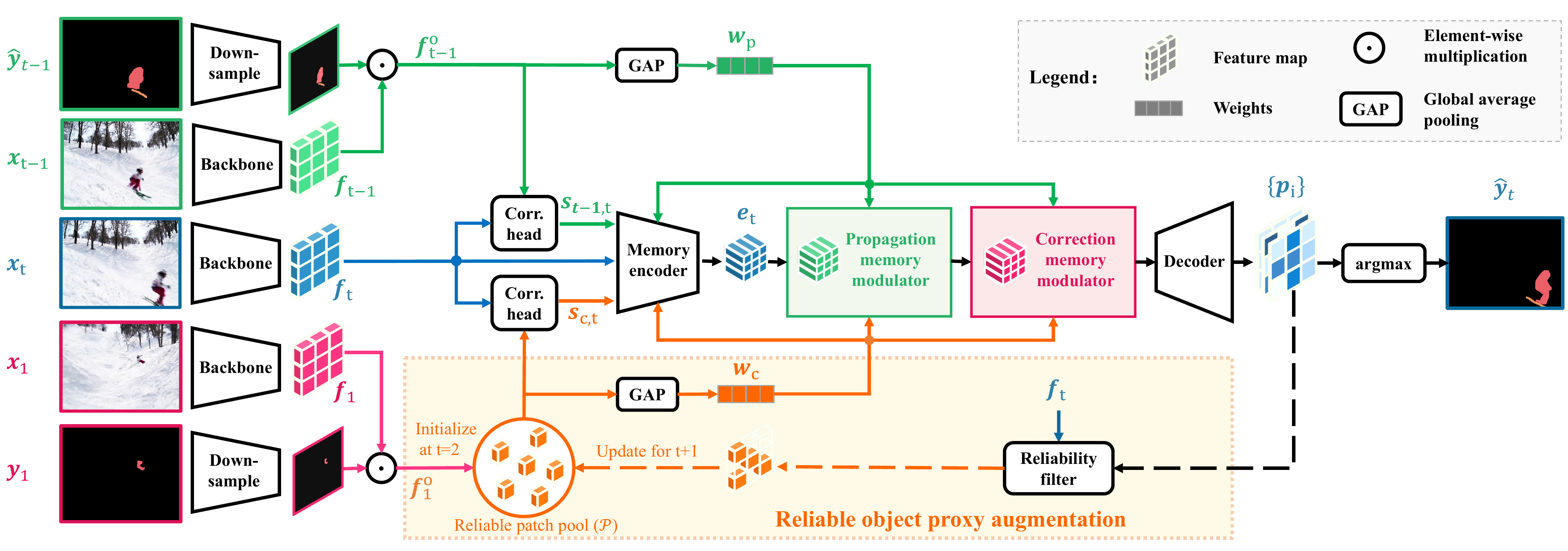} 
	\caption{Overview of the proposed framework. We disentangle the correction mechanism from the frame-to-frame mask propagation process. 
	We assemble a cascaded scheme of propagation-correction modulators to leverage local temporal correlations and reliable references in order. We also augment the reference cues by supplementing reliable feature patches to a maintained pool, thus offering more comprehensive and expressive object representations to the modulators. A reliability filter is introduced to filter out uncertain patches for subsequent frames.
	}
	\label{fig:PIPELINE}
\end{figure*}

\section{Related Work}

\paragraph{Propagation-based VOS.}
Propagation-based methods utilize the semantic or spatial cues from the previous frame to predict the mask of the current frame.	
Early methods \cite{perazzi2017learning,caelles2017one,NIPS2017_6c9882bb,khoreva2019lucid} utilize online-learning method to eliminate the drifting problem but is time-consuming. 
Optical flow \cite{tsai2016video,NIPS2017_6c9882bb,cheng2017segflow,xu2018dynamic} and object tracking also prove to be useful guidance for mask propagation.
Although propagation-based models can secure good temporal consistency\cite{caelles2017one}, these propagation-based methods are prone to error accumulation, which may largely degrade the VOS performance especially for long video clips\cite{liang2020video}. 

\paragraph{Matching-based VOS.}
Matching-based methods learn an embedding space for target objects. \cite{chen2018blazingly,hu2018videomatch,zeng2019dmm} directly construct the correspondence between the current frame with the first frame. \cite{lin2019agss,voigtlaender2019feelvos,wang2019ranet,yang2020collaborative} further leverage both the first and the previous frames. Several recent methods \cite{hu2018videomatch,liang2020waternet,duke2021sstvos} turn to use several latest frames to further improve the local temporal guidance. Moreover, STM-based networks \cite{oh2019video,seong2020kernelized,lu2020video,liang2020video,NEURIPS2020_liangVOS,cheng2021modular,wang2021swiftnet,xie2021efficient,hu2021learning,seong2021hierarchical} boost the performance with memory networks that memorize information from past frames for further reuse, which relieve the error propagation to some extent. However, how to reduce uncertainty propagation is still a hard problem and hasn't been tackled perfectly. Our method disentangles the guidance by reliability and further suppresses uncertainty with a carefully designed scheme. 
%, with an elaborate training procedure using extensive simulated data
\paragraph{Conditional modulation.}
Recently, \cite{Yang2018osmn,yang2020collaborative,li2021hybrid} introduce conditional modulation methods to tackle video-related tasks like video object segmentation for instance-based, or spatial guidance. However, the modulation weights or embeddings for mask propagation in the video are inevitably volunteered to error. %different from style transfer tasks where the weights to be conditioned on are relatively error-free with human interference,
To collaboratively suppress error propagation and make full use of the power of conditional modulation, we propose a new conditional modulation method called \textit{reliable conditional modulation} to tackle VOS.

\section{Preliminaries}
We first review the probabilistic model of frame-to-frame VOS and analyze it from two aspects, \ie, the frame-by-frame propagation path and the correction path from reliable reference. We also introduce a widely used measurement for prediction reliability.

\subsection{Probabilistic model of frame-to-frame VOS}
Given all available observations $\mathbf{x}_{1:t}$ up to $t$-th frame, the label up to t-th frame $\mathbf{y}_{1:t}$ is predicted by maximum a posterior (MAP) estimation:

\begin{small}
    \begin{equation}
    \label{eq:map}
    p(\mathbf{y}_{1:t} | \mathbf{x}_{1:t}) = \frac{p(\mathbf{x}_{1:t} | \mathbf{y}_{1:t}) p(\mathbf{y}_{t})}{p(\mathbf{x}_{1:t})} \propto p(\mathbf{x}_{1:t} | \mathbf{y}_{1:t}) p(\mathbf{y}_{t})
    \end{equation}
\end{small}

Here $p(\mathbf{x}_{1:t}|\mathbf{y}_{1:t})$ is the observation model, which is usually estimated by the likelihood $p(\mathbf{x}_{1:t}|\mathcal{D})$, where $\mathcal{D}$ denotes the training data. 
$p(\mathbf{y}_{1:t-1}|\mathbf{x}_{1:t-1})$ is the posterior up to previous frame.
$p(\mathbf{y}_{1:t})$ is the prior model and could be unfolded with the first-order markov assumptions:
\begin{small}
    \begin{eqnarray}
    \label{eq:prior}
        p(\mathbf{y}_{1:t}) = p(\mathbf{y}_{t} | \mathbf{y}_{t-1}) p(\mathbf{y}_{1:t-1})
        = p(\mathbf{y}_{1}) \Pi_{2}^t p(\mathbf{y}_{i} | \mathbf{y}_{i-1})
    \end{eqnarray}
\end{small}
We then instantiate \eqn{\ref{eq:map}} with the propagation and correction path respectively.
\paragraph{Propagation path.}
We assume that the observation model is conditionally independent given the states, \ie, $p(\mathbf{x}_{1:t}|\mathbf{y}_{1:t}) = \Pi_{1}^t p(\mathbf{x}_{i}|\mathbf{y}_{i})$. The posterior takes the form
\begin{small}
\begin{eqnarray}
\label{eq:prob_propation}
p(\mathbf{y}_{1:t} | \mathbf{x}_{1:t}) & \propto & \Pi_{1}^t p(\mathbf{x}_i|\mathbf{y}_i) \Pi_{2}^t p(\mathbf{y}_i|\mathbf{y}_{i-1}) p(\mathbf{y}_1) \nonumber \\ & \propto & \Pi_{1}^t p(\mathbf{x}_i|\mathbf{y}_i) \Pi_{2}^t p(\mathbf{y}_i|\mathbf{y}_{i-1})
\end{eqnarray}
\end{small}
Note $p(\mathbf{y}_1)$ is omitted since the label of the first frame $\mathbf{y}_1$ is given.
Therefore, we observe that prediction uncertainty of $p(\mathbf{y}_t|\mathbf{y}_{t-1})$ will accumulate over time, which lead to error propagation.

\paragraph{Correction path.}
We first consider the direct translation from the reliable reference frame $\mathbf{x}_1$ to the $t$-th frame $\mathbf{x}_t$ for correction. The posterior takes the form
\begin{small}
\begin{eqnarray}
\label{eq:prob_match}
p(\mathbf{y}_1, \mathbf{y}_t | \mathbf{x}_{1}, \mathbf{x}_{t}) & = & \frac{p(\mathbf{y}_t, \mathbf{x}_t | \mathbf{y}_{1}, \mathbf{x}_{1}) p(\mathbf{x}_1, \mathbf{y}_1)}{p(\mathbf{x}_1, \mathbf{x}_t)} \nonumber \\ & \propto & p(\mathbf{y}_t, \mathbf{x}_t | \mathbf{y}_{1}, \mathbf{x}_{1})
\end{eqnarray}
\end{small}
Again, $p(\mathbf{x}_1, \mathbf{y}_1)$ is omitted since $\mathbf{x}_1$ and $\mathbf{y}_1$ are given. The joint condition probability $p(\mathbf{y}_t, \mathbf{x}_t | \mathbf{y}_{1}, \mathbf{x}_{1})$ corresponds to the joint similarity of observations and labels between target and reference frames. Since labels represent object masks, the joint similarity can be considered as an object-aware similarity. Thus, prediction in the correction path is highly correlated to the object-aware similarity between target and reference frames. Since reference in the first frame may not be comprehensive, we will augment it during the iterations.

\paragraph{Prediction reliability.}
Prediction uncertainty of deep neural networks is difficult to estimate accurately, while it is highly correlated with information entropy \cite{272494}. Here, we employ the Shannon entropy to measure the reliability of prediction in each iteration:
\begin{small}
\begin{eqnarray}
\label{eq:shannon_entropy}
H(I) &= -\sum_{i=1}^{N+1} P (\frac{e^{\mathbf{p}_{i}}}{\sum_{j=1}^{N+1} e^{\mathbf{p}_{j}}}) \log P (\frac{e^{\mathbf{p}_{i}}}{\sum_{j=1}^{N+1} e^{\mathbf{p}_{j}}})
\end{eqnarray}
\end{small}
where $\mathbf{p}_{i}, i \in \{1, ..., N+1\}$ indicate the probability maps of $N$ objects and background.

\section{Reliable Propagation-Correction Modulation}\label{sec:modulation}
In this section, we first describe the overview of our pipeline and then elaborate on the reliable modulation mechanism. Finally, the network design is detailed.

%\subsection{Network Modulation for VOS}
\subsection{Pipeline Overview}
\fig{\ref{fig:PIPELINE}} illustrates the overview of our framework. Our goal is to predict the object mask $\hat{\mathbf{y}}_t$ of the target frame $\mathbf{x}_t$, given the first frame $\mathbf{x}_1$ and its object mask $\mathbf{y}_1$, as well as the previous frame $\mathbf{x}_{t-1}$ and predicted object mask $\hat{\mathbf{y}}_{t-1}$.

We first extract features maps $\mathbf{f}_1, \mathbf{f}_{t-1}, \mathbf{f}_t$ of frames $\mathbf{x}_1, \mathbf{x}_{t-1}, \mathbf{x}_t$ from a shared backbone respectively.
To make features object-aware, we mask the features with corresponding object masks, $\mathit{f}_{t'}^{o} = \mathbf{f}_{t'}$ $\odot$ $\mathbf{y}_{t'}$, $t'=\{1,t-1\}$.
For the propagation path, we define an object proxy $w_{p}$ representing a image-level object feature by applying a global average pooling (GAP) on $\mathit{f}_{t-1}^o$. Object-aware inter-frame correlation for propagation path is then represented with a similarity map $\mathbf{s}_{t-1, t}$ between the target feature $\mathbf{f_t}$ and masked previous feature $\mathit{f}_{t-1}^o$.
For the correction path, we also define an object proxy $w_{c}$ and a similarity map $\mathbf{s}_{c, t}$; instead of directly computing them from a reference frame feature, we maintain a reliable feature patch pool containing reliable object feature patches from historical frames.
The object proxy and correlation cues from both paths are then further encoded with a memory encoder to form a compact embedding $\mathbf{e}_t$ for the target frame. Given two object proxies $w_{p}$ and $w_c$, the current embedding is modulated with the cascaded propagation and correction modulator respectively. 
The final probability $\mathbf{p}$ for each object is decoded from the modulated embedding with a decoder. 
In turn, we provide the reliability map $\mathbf{r}_t$ for updating the reliable patch pool $\mathcal{P}$ for the next frame with the reliability filter.

\begin{figure}[t!]
		\centering
		\includegraphics[width=0.45\textwidth]{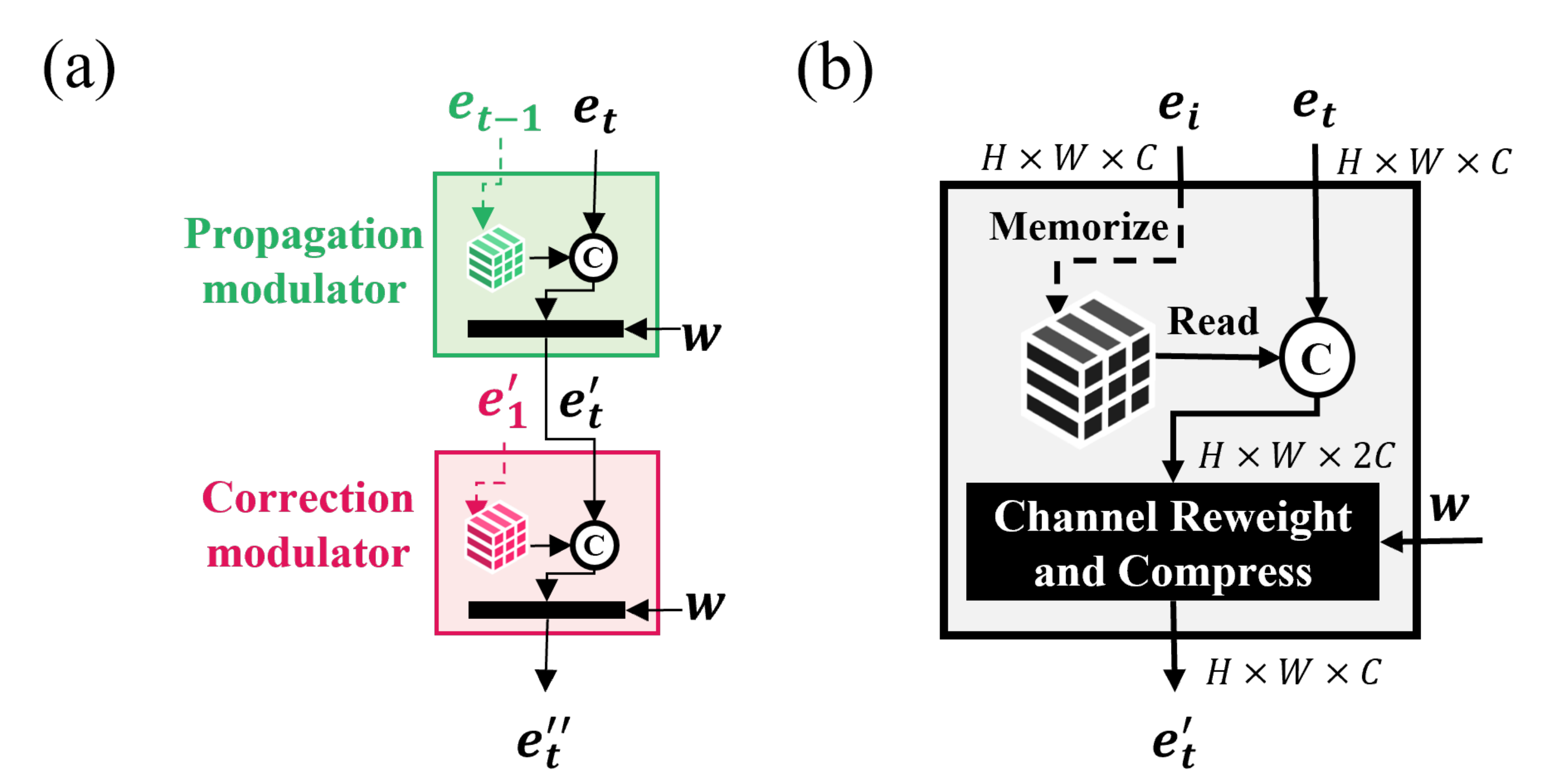}
		\caption{(a) Cascaded propagation-correction ($P2C$) modulator scheme. (b) Basic modulator block.}
		\label{fig:Mem_modu_simplified} 
\end{figure}

\subsection{Object proxy for Correction Path}
\paragraph{Reliable object patch pool.}
For correction path, the object semantics in the first frame is often incomplete. Specifically, the mask may cover only a part of an object. This will highly degrade the correction effects from the reference frame.
Thus, we need to augment the object proxy by supplementing new features in the process of mask propagation. However, previous study \cite{wu2020memory} has shown that mask propagation with multiple historical frames may be vulnerable to the influence of inaccurate information, leading to error propagation. 
To augment object proxy with historical cues while suppressing error propagation, we introduce a \textit{reliable object patch pool} for updating useful feature patches for object representation augmentation and a \textit{reliability filter} to filter out uncertain feature patches.

Specifically, we use a reliability map $\mathbf{r}$ to filter out uncertain patches, i.e., $\mathbf{f_c} = \mathbf{f} \odot \mathbf{y} \odot \mathbf{r}$. During mask propagation, we only recall the highly reliable patches of the object and supplement them to the reliable object patch pool $\mathcal{P}$, thus augmenting the semantic cues of target objects. Details are described in Algorithm \ref{alg:rop}.

\paragraph{Reliable object proxy.}
Given the reliable object patch pool $\mathcal{P}$ containing a set of highly reliable patches, we construct the reliable object proxy $w_{c}$ by apply GAP on all elements in $\mathcal{P}$, i.e.,  $w_c=\text{GAP}(\{\mathbf{f} \odot \mathbf{y} \odot \mathbf{r}\}, \mathbf{f} \in \mathcal{P})$.

\begin{algorithm}[tb]
\caption{Reliable object proxy augmentation}
\label{alg:rop}
\textbf{Input}: Embedding of current frame $\mathbf{f}_{t}$, embedding $\mathbf{f}_{t-1}$, mask $\hat{\mathbf{y}}_{t-1}$ and reliability map $\mathbf{r}_{t-1}$ of previous frame, reliable object patch pool $\mathcal{P}$ and its updating time interval $\tau$, timestamp $t \in \{2, ...\}$. \\
\textbf{Output}: Augmented reliable object proxy $w_c$ and similarity map from the correction path $s_{c,t}$ 
\begin{algorithmic}[1] %[1] enables line numbers
\IF{$\mathbf{t}==2$}
\STATE Let $\mathbf{r}_1 = \mathbf{I}$, $\mathcal{P} =  \varnothing$. 
\ENDIF
\IF{($({t-2}) \mod {\tau}) == 0$}
\STATE $\mathit{f}_{t-1}^o = \mathbf{f}_{t-1} \odot \hat{\mathbf{y}}_{t-1} \odot \mathbf{r}_{t-1}$ 
\STATE $\mathcal{P}=\mathcal{P} \bigcup \mathsf{f}_{t-1}^o$. 
\ENDIF
\STATE $w_c = \frac{\sum_{f \in \mathcal{P} } f }{|\mathcal{P}|} $
\STATE $s_{c,t} = \max \text{S}({\mathcal{P},\mathbf{f}_{t}})$, \\where $\text{S}$ is a similarity measure.
\STATE \textbf{return} $w_c, s_{c,t}$.
\end{algorithmic}
\end{algorithm}

\subsection{Propagation-Correction Modulation}

While most methods \cite{oh2019video,seong2020kernelized,lu2020video} employ reference and previous cues equivalently, they do not distinguish their influence on the prediction of the current frame. We address that mask propagation along the propagation path can preserve local temporal consistency, while the reference provides more reliable information. The reliable reference is more suitable to perform a correction role after propagation, and thus such two kinds of cues should be handle in a disentangled way to be made full use of.

We maintain two types of external memory modules to selectively memorize information from the propagation and correction paths, namely, \textit{propagation memory} storing the previous embedding $\mathbf{e}_{t-1}$ and updating at each frame, and \textit{correction memory} storing the embedding of reference frame $\mathbf{e}_{1}$.
Accordingly, we build two modulator $\phi_{prop}$ and $\phi_{corr}$ to modulate the current embedding with corresponding object proxy $w_{p}$ and $w_c$.
Intuitively, we set the correlation modulator after the propagation memory as illustrated in \fig{\ref{fig:Mem_modu_simplified}} (a) because correlation is reliable and should not be overridden by the propagation modulator.

Our key observation is that the order of modulators in cascaded schemes matters since the uncertainty also relies on the depth of relevant layers in the neural network~\cite{goldfeld2019estimating}.
We verified this observation with detailed analysis in the experiment section and confirmed a cascade with propagation-correction order (i.e. $P2C$) outperforms other cascade variants as well as parallel approaches.
\begin{figure*}[ht!]
	\centering
	\includegraphics[width=\textwidth]{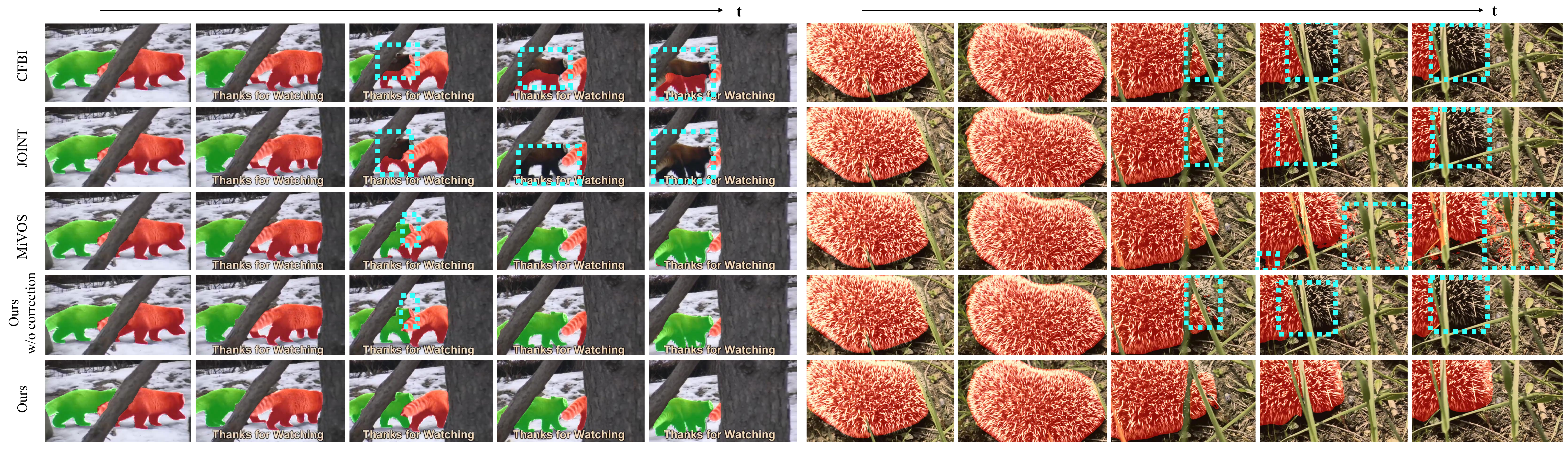}
\caption{Qualitative comparison to several state-of-the-art methods, CFBI~\cite{yang2020collaborative}, JOINT~\cite{mao2021joint}, MiVOS~\cite{cheng2021mivos} on YouTube-VOS 19 validation set. With the reliable correction mechanism, our model can reduce the error regions in the mask propagation process. Error regions are highlighted with blue bounding boxes.}
	\label{fig:qualitative_comparison}
\end{figure*}
\subsection{Network Design}
We detail the implementation of each network module.
\paragraph{Correlation head.}
The correlation head for calculating similarities between features consists of a set of linear operations in local windows as \cite{voigtlaender2019feelvos}, and then the similarity maps are concatenated with the previous mask and projected into a high-dimensional space. Note that we follow \cite{yang2020collaborative} to split features into the foreground and background-masked ones according to corresponding masks and concatenated them together for further processing.

\paragraph{Memory encoder.}
The memory encoder $\phi_{me}$ maps the concatenated features $[\mathbf{f}_t, \mathbf{s}_{c,t}, \mathbf{s}_{t-1,t}]$ to a lower dimension and compact embedding $\mathbf{e}_t$. Meanwhile, it consists of a $1 \times 1$ convolution layer followed by a series of channel re-weighting operations. $\mathbf{e}_t=\phi_{me}([\mathbf{f}_t;\mathbf{s}_{c,t};\mathbf{s}_{t-1, t}],[w_{p}; w_c])\label{eq:memory_encoder1}$, where $[;]$ denotes concatenation.

\paragraph{Modulator block.}

\fig{\ref{fig:Mem_modu_simplified}} (b) illustrates the structure of a basic memory modulator block.
A memory modulator block maintains a memory buffer $\mathbf{e}_{i} \in\mathbb{R}^{H \times W \times C}$ for later usage. During each forward, the memory modulator block inputs memory embedding features $\mathbf{e}_{t}$ from the current frame, reads out the buffered memory embedding (i.e., $\mathbf{e}_{t-1}$ for propagation modulator or $\mathbf{e}_1$ for correction modulator), concatenates them together and then performs a reweighting operation along channel dimension with $w_{p}$ or $w_c$. 

Inspired by the theoretical proof \cite{chen2011lowrank} that low-rank matrix can recover from errors and erasures, we implicitly force the memory embedding to simultaneously keep the low-rank property and encode more useful information by compressing the reweighted embedding to a lower dimension via a $1 \times 1$ convolution layer.

\paragraph{Reliability filter.}
We first compute the Shannon entropy $H_t$ from  probability maps $\{\mathbf{p}_{i}\}$ with \eqn{\ref{eq:shannon_entropy}}. Then the reliability map $\mathbf{r}_t$ is estimated by applying a threshold function ${\psi}_{\alpha}(\cdot)$ on the Shannon entropy $H_t$ to remain the reliable regions in the final mask prediction for the update of the reliable patch pool.

\section{Experiment}\label{sec:Experiment}

\paragraph{Datasets.} We evaluate our model mainly on two widely used VOS benchmarks with multiple objects, YouTube-VOS \cite{xu2018youtube} and DAVIS17 \cite{pont20172017}, and a small-scale single object VOS benchmark DAVIS16 \cite{perazzi2016benchmark}.
The unseen object categories make YouTube-VOS more proper to measure the generalization ability of algorithms. So we conduct our experiments on YouTube-VOS to evaluate various methods accurately. 

\paragraph{Metrics.} We adopt the evaluation metrics from the DAVIS benchmark \cite{perazzi2016benchmark}: the region accuracy J, which calculates the intersection-over-union (IoU) of the estimated masks and the ground truth masks, the boundary accuracy F, which measures the accuracy of boundaries via bipartite matching between the boundary pixels. 

\paragraph{Implementation details.}

We use the DeepLabv3+ \cite{chen2018encoder} architecture as the backbone of our network. 
Unless otherwise specified, we use ResNet-101 as the backbone network of DeepLabv3+.

The training is conducted with an SGD optimizer with a momentum of 0.9 using the cross-entropy loss.
For YouTube-VOS experiments, we only use YouTube-VOS without any external datasets. We first use a learning rate of 0.02 for 200$k$ steps with a batch size of 8, then change to a learning rate of 0.01 for another 200$k$ steps. 
During inference, we restrict the long-edge of each frame to no more than 1040 pixels and apply a scale set of [1.0, 1.3, 1.5] for multi-scale testing on YouTube-VOS. For DAVIS17 and DAVIS16 experiments, we finetune the trained model on YouTube-VOS for 20$k$ steps with both DAVIS and YouTube-VOS in a ratio of 2:1. 

All the experiments are performed on an NVIDIA DGX-1 Linux workstation (OS: Ubuntu 16.04.4 LTS, GPU: 8$\times$ Tesla V100). Our code is implemented with PyTorch 1.4.1 and is partly leveraged from~\cite{CFBI_REPO}. To ensure the validity of experiments, our main results are averaged for 3 runs. For hyper-parameters, we set the update interval $\tau$ for reliable patch candidate pool $\mathcal{P}$ as 5 and the parameter $\alpha$ in the reliability filter as 1 without tuning.

\begin{table*}[ht]
\caption{Quantitative comparisons on YouTube-VOS. Subscript $s$ and $u$ denote scores in seen and unseen categories. $\bigtriangleup$ denotes using external training datasets. Superscript $MS$ and $F$ denotes using multi-scale and flip testing in evaluation respectively. } 
\centering
\setlength{\tabcolsep}{3.5mm}
\resizebox{\textwidth}{!}
    {

	\begin{tabular}{llllll|lllll}%{width=0.95\textwidth}
		\toprule
		\multirow{2}{*}{Methods} &\multicolumn{5}{c}{YouTube-VOS 2018 Validation}  & \multicolumn{5}{c}{YouTube-VOS 2019 Validation}                                    \\
		\cmidrule(l){2-11}
		 & J$\&$F & J${}_{s}$ & J${}_{u}$ & F${}_{s}$ & F${}_{u}$ & J$\&$F & J${}_{s}$ & J${}_{u}$ & F${}_{s}$ & F${}_{u}$\\
 \midrule 
		AGAME {\cite{johnander2019generative}}&      66.1 & 67.8 & 60.8 & -     & -   & -     & -& -    &-&- \\
		PReM {\cite{luiten2018premvos}}&    66.9 & 71.4 & 56.5 & 75.9  & 63.7 & -     & -& -     & -  & -   \\

		STM${}^{\bigtriangleup}$ {\cite{oh2019video}}&      79.4 & 79.7 & 72.8 & 84.2  & 80.9 & -     & -& -     & -  & -   \\
		CFBI {\cite{yang2020collaborative}}&    81.4 & 81.1 & 75.3 & 85.8  & 83.4  &81.0 & 80.6 & 75.2 & 85.1  & 83.0\\		
		RMN {\cite{xie2021efficient}}&         81.5&82.1&75.7& 85.7&82.4 & -     & -& -     & -  & -    \\
		SST {\cite{duke2021sstvos}}&       81.7&81.2&76.0& -&-  & 81.8&80.9&76.6     & -  & -    \\
		LCM${}^{\bigtriangleup}$ {\cite{hu2021learning}}&  82.0&82.2&75.7& {86.7}&83.4 & -     & -& -     & -    & -  \\	
		MiVOS+km${}^{\bigtriangleup}$ {\cite{cheng2021modular}}&  82.6 & 81.1 &77.7&85.6&86.2 &  82.8 & 81.6 &77.7&85.8&85.9       \\
        HMMN${}^{\bigtriangleup}${\cite{seong2021hierarchical}}&   82.6 & 82.1& 76.8&87.0&84.6 & 82.5 & 81.7 &77.3&86.1&85.0     \\
		CFBI+ {\cite{yang2021collaborative}}&   82.8 & 81.8 &77.1&86.6&85.6 & 82.9 & 80.6 &78.9&85.2&86.8     \\

        JOINT {\cite{mao2021joint}}&    83.1 & 81.5 &\textbf{78.7}&85.9&86.5 & 82.8 & 80.8 &79.0&84.8&86.6     \\
		\textbf{Ours}        &   \textcolor{black}{\textbf{{84.0}}} & \textcolor{black}{\textbf{83.1}} & \textcolor{black}{{78.5}} & \textcolor{black}{\textbf{87.7}}  & \textcolor{black}{\textbf{86.7}} &   \textcolor{black}{\textbf{83.9}} & \textcolor{black}{\textbf{82.6}} & \textcolor{black}{\textbf{79.1}} & \textcolor{black}{\textbf{86.9}}  & \textcolor{black}{\textbf{87.1}}\\
        \midrule
	    CFBI${}^{MS,F}$ {\cite{yang2020collaborative}}&    82.7 & 82.2  & 76.9 & 86.8  & 85.0  &      82.4 & 81.8 & 76.9 & 86.1  & 84.8   \\	
		\textbf{Ours${}^{MS}$} &\textbf{84.3} & \textbf{83.3}&\textbf{78.9}&\textbf{87.9} &\textbf{86.9}  & \textbf{84.2} & \textbf{83.0}  & \textbf{79.4} &\textbf{87.3}  &    \textbf{87.2}  \\	
		\bottomrule
	\end{tabular}

    }
	\label{table:ytb}
\end{table*}

\begin{table}
	\centering
	\caption{Quantitative comparisons on DAVIS. $\bigtriangleup$ denotes using external training datasets besides YouTubeVOS and DAVIS. Superscript $FR$ denotes full-resolution testing. Otherwise, methods are all tested on $480p$. }
	\setlength{\tabcolsep}{1.0mm}
	\begin{small}
	\resizebox{0.48\textwidth}{!}
	{
    \begin{tabular}{c|ccccccc|c}
    \toprule
           & CFBI & SST & MiVOS${}^{\bigtriangleup}$ & RMN${}^{\bigtriangleup}$ & LCM  & JOINT & \textbf{Ours} & \textbf{Ours${}^{FR}$}\\ \midrule
         \multicolumn{9}{c}{DAVIS16 Validation (single object, easy)}                                    \\
		\midrule
        J$\&$F   & 89.4   & - & \textbf{91.0}      & 88.8& 90.7 & - & 90.6 & \textbf{91.5}  \\ 
        J     & 88.3    & - & {89.7} & 88.9 &\textbf{89.9} & - & 87.1 & \textbf{88.3}  \\ 
        F    & 90.5    & - & 92.1         & 88.7  &91.4& - &\textbf{94.0}& \textbf{94.7} \\ 
        \midrule
         \multicolumn{9}{c}{DAVIS17 Validation (multi-object, medium)}                                    \\
		\midrule
        J$\&$F   & 81.9   & 82.5 & 83.3 & 83.5 & 83.5 & 83.5 & \textbf{83.7} & \textbf{84.8}  \\ 
        J     & 79.1    & 79.9 & 80.6 & 81.0 & 80.5 & 80.8 & \textbf{81.3} & \textbf{82.5}  \\ 
        F    & 84.6    & 85.1 & 85.1 & 86.0 & \textbf{86.5} & 86.2 & {86.0} & \textbf{87.2} \\ 
        \midrule
         \multicolumn{9}{c}{DAVIS17 Test-dev (multi-object, hard)}                                    \\
		\midrule
        J$\&$F   & 74.8     & -  & 76.5 & 75.0  & 78.1 & - & \textbf{79.2} & \textbf{81.0} \\ 
        J     & 71.1     & - & 72.7  & 71.9 & 74.4 & - & \textbf{75.8}  &  \textbf{77.6}\\ 
        F     & 78.5    & - & 80.2  & 78.1 & 81.8 & - & \textbf{82.6}  & \textbf{84.3}\\         
        
        \bottomrule
    \end{tabular}
    }
    \end{small}
    \label{table:dv}
\end{table}
\subsection{Main Results}
\paragraph{Quantitative comparison.}
We compare our method with multiple state-of-the-art methods on YouTube-VOS18 validation (YV18-Val), YouTube-VOS19 validation (YV19-Val) and DAVIS benchmarks in Table~\ref{table:ytb} and Table~\ref{table:dv}. Without using any bells and whistles (e.g., fine-tuning at test time, top-k filtering, pre-training on external training dataset BL30K  or simulated training data), our model significantly outperforms nearly all the contemporary work and previous SOTA methods. Our model stands out in most of the evaluation metrics, especially on unseen categories of YouTube-VOS, which further demonstrates the generalization ability. On the challenging DAVIS17 Test-dev split, the overall performance can be further promoted to 81\% J\&F with full-resolution testing thanks to the good scalability of input resolution of our model.

\paragraph{Qualitative comparison.}
 \fig{\ref{fig:qualitative_comparison}} shows the qualitative comparison between state-of-the-art methods and our model on the YouTube-VOS validation set. Thanks to our reliable correction mechanism, our model suppresses the error propagation better, achieving better results when the targets become different from the reference.

\begin{table}[tbp]
	\centering
	\caption{Ablation on memory modulator scheme variants on YouTube-VOS 18 validation set.  $\uparrow$ indicates improvement over our compared method CFBI.  }
	\setlength{\tabcolsep}{0.9mm}
	\resizebox{0.48\textwidth}{!}
	{\begin{small}
	\begin{tabular}{@{}lllllll@{}}
		\toprule
		\multicolumn{2}{c}{Method}                        & J$\&$F  & J${}_{s}$ & J${}_{u}$ & F${}_{s}$ & F${}_{u}$   \\ \midrule
		\multicolumn{2}{c}{\multirow{1}{*}{CFBI}} &81.4 & 81.1 & 75.3 & 85.8  & 83.4  \\ 
		\midrule 
		
		\multirow{7}{*}{\begin{tabular}[c]{@{}l@{}}  \\ \\ Ours \end{tabular}} & S2S    & 81.9\scriptsize{(0.5$\uparrow$)}&81.6\scriptsize{(0.5$\uparrow$)}&76.1\scriptsize{(0.8$\uparrow$)}&86.1\scriptsize{(0.3$\uparrow$)}&83.9\scriptsize{(0.5$\uparrow$)} \\  %65.552664
		\cmidrule(l){2-2}
		& $S2P$   & 81.9\scriptsize{(0.5$\uparrow$)} &81.4\scriptsize{(0.3$\uparrow$)}&76.2\scriptsize{(0.9$\uparrow$)}&85.9\scriptsize{(0.1$\uparrow$)}&84.3\scriptsize{(0.9$\uparrow$)} \\
		& $P2S$    & 82.2\scriptsize{(0.8$\uparrow$)} &81.8\scriptsize{(0.7$\uparrow$)}&76.4\scriptsize{(1.1$\uparrow$)}&86.5\scriptsize{(0.7$\uparrow$)}&84.3\scriptsize{(0.9$\uparrow$)} \\
		& $P2P$    & 82.3\scriptsize{(0.9$\uparrow$)} &81.8\scriptsize{(0.7$\uparrow$)}&76.3\scriptsize{(1.0$\uparrow$)}&86.5\scriptsize{(0.7$\uparrow$)}&84.4\scriptsize{(1.0$\uparrow$)} \\
		\cmidrule(l){2-2}
        & $C2S$    & 82.1\scriptsize{(0.7$\uparrow$)} &81.8\scriptsize{(0.7$\uparrow$)}&{{76.2\scriptsize{(0.9$\uparrow$)}}}&86.3\scriptsize{(0.5$\uparrow$)}&{{84.1\scriptsize{(0.7$\uparrow$)}}}  \\
		& $S2C$    & 82.6\scriptsize{(1.2$\uparrow$)} &81.7\scriptsize{(0.6$\uparrow$)}&\textbf{{77.5\scriptsize{(2.2$\uparrow$)}}}&86.1\scriptsize{(0.3$\uparrow$)}&\textbf{{85.2\scriptsize{(1.8$\uparrow$)}}}  \\
		& $C2C$    & 82.3\scriptsize{(0.9$\uparrow$)} &81.6\scriptsize{(0.5$\uparrow$)}&{76.8\scriptsize{(1.5$\uparrow$)}}&86.1\scriptsize{(0.3$\uparrow$)}&{84.7\scriptsize{(1.3$\uparrow$)}}\\
		\cmidrule(l){2-2}
		& $C2P$     & 82.5\scriptsize{(1.1$\uparrow$)} &82.0\scriptsize{(0.9$\uparrow$)}&76.7\scriptsize{(1.4$\uparrow$)}&86.7\scriptsize{(0.9$\uparrow$)}&84.4\scriptsize{(1.0$\uparrow$)} \\ 
		& $P2C$     & \textbf{82.9\scriptsize{(1.5$\uparrow$)}}  &\textbf{82.9\scriptsize{(1.8$\uparrow$)}}&76.9\scriptsize{(1.6$\uparrow$)}&\textbf{87.4\scriptsize{(1.6$\uparrow$)}}&{84.5\scriptsize{(1.1$\uparrow$)}}\\		
		& $P\&C$    & 82.5\scriptsize{(1.1$\uparrow$)} &82.1\scriptsize{(1.0$\uparrow$)}&76.7\scriptsize{(1.4$\uparrow$)}&86.7\scriptsize{(0.9$\uparrow$)}&84.3\scriptsize{(0.9$\uparrow$)}\\ 
        \cmidrule(l){2-2}
		& $P2P2P$    & 82.3\scriptsize{(0.9$\uparrow$)} &82.0\scriptsize{(0.9$\uparrow$)}&76.4\scriptsize{(0.9$\uparrow$)}&86.5\scriptsize{(0.7$\uparrow$)}&84.1\scriptsize{(0.7$\uparrow$)} \\

		\bottomrule
	\end{tabular}
	\end{small} 
}
	\label{table:4}
\end{table}

\begin{figure}[t]
		\centering
		\includegraphics[width=0.48\textwidth]{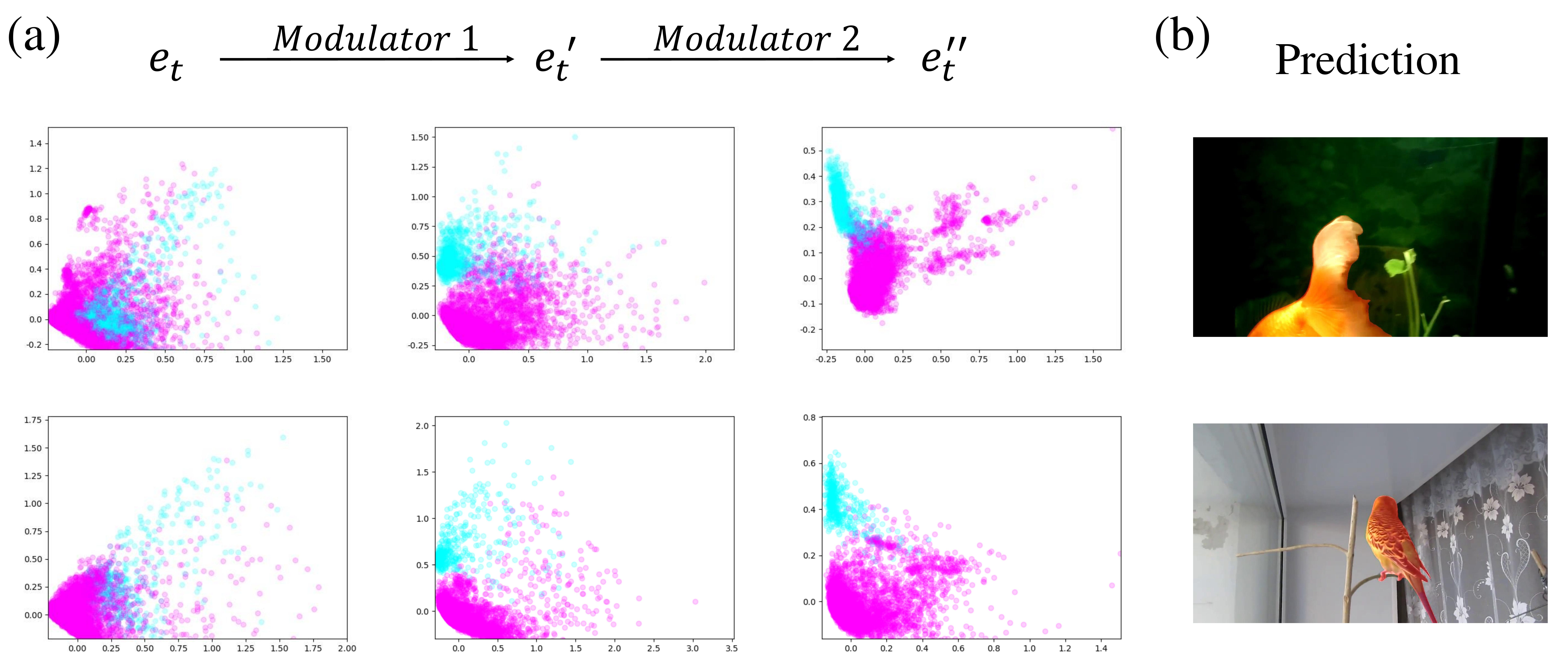}%,height=3.00cm
		\caption{(a) Pixel-level embedding evolution through a $P2C$ modulator. Features of foreground and background are colored in blue and purple. The visualization is conducted by reducing the embedding to 2-dims with PCA. (b) Predicted mask with input frame blended. Zoom in to view better.}
		\label{fig:pca_embedding} 
\end{figure}

\subsection{Ablation Study and Discussion}
\label{sec:ablation_study}
In addition to the state-of-the-art performance, we provide the following insights with our detailed ablation.

\paragraph{Modulator assembly variants.}
Apart from propagation modulator $P$ and correction modulator $C$, an auxiliary self-modulator $S$ that uses current embedding both as both memory and input serves as a reference. For simplicity, "2" and "\&" between characters stand for cascaded and parallel assembly schemes. The variants can be categorized into propagation-based ($S2P$, $P2S$, $P2P$ and $P2P2P$), correction-based ($S2C$, $C2S$ and $C2C$) and propagation~\&~correction ($C2P$, $P2C$ and $P\&C$) schemes. Note that $P\&C$ is a variant where the propagation and correction modulators are assembled in parallel streams and followed by a self-modulator to fuse the outputs.

\paragraph{Modulation with propagation and correction guidance.}
To study how the propagation and correction guidance influences VOS results, we conduct ablation experiments on the YV18-Val split with different modulator variants.
We also list the SOTA model CFBI that uses naive local and global guidance for comparison.
The results are shown in \tab{\ref{table:4}}.

\textbf{\textit{Can our modulator improve the performance only with direct propagation guidance?}} Yes. From Table \ref{table:4}, we can find that our propagation-based variants all outperform CFBI. The reasons are two-fold. First, while CFBI utilizes image-level features from previous frames for improving performance, it does not have explicit designs to suppress error propagation. Instead, our model utilizes a compressed and memorized embedding throughout the whole sequence, which consolidates the local spatial correlation. Secondly, the modulator is a plug-and-play module with a resolution-keeping design, which may help protect the detail from loss.

\textbf{\textit{Can more propagation guidance from different layers further boost the performance?}} No. Although modulators are light-weight, simply stacking more propagation-based modulators ($P2S$, $P2P$ and $P2P2P$) only has a marginal gain, which indicates that simply incorporating propagation guidance is not enough.

\textbf{\textit{Is correction guidance effective for uncertainty suppression?}} Yes. Thanks to the highly reliable correction memory, simply leveraging the correction modulator ($C2S$, $S2C$, $C2C$) can boost the performance especially in unseen categories compared with the one without using correction guidance ($S2S$). 
Among those injections, assembling the correction modulator in deeper layers ($S2C$) brings the largest gain since it encodes cues with high reliability and its correction impact will not be overridden by others.

\textbf{\textit{Can propagation and correction be collaboratively leveraged?}} Yes. The propagation-correction cascaded assembly ($P2C$) stands out among all the modulator variants, which verifies our insight that propagation modulator relying on local coherence fits shallow layers while correction modulator relying on high-level semantics fits deeper layers. 

\begin{table}[ht]

\caption{Ablation study for reliable object proxy augmentation on YouTube-VOS 19 validation set. $P2C$ denotes the propagation-correction modulator assembly, $OA$ and $RF$ denotes using object proxy augmentation and reliability filter. $W$ denotes using $OA$ for calculate modulation weight only. $W+S$ denotes using $OA$ for calculating weights and correlation (similarity).}
	\setlength{\tabcolsep}{3mm}
	\resizebox{0.48\textwidth}{!}
	{
\begin{tabular}{@{}lll|lllll@{}}
\toprule

$P2C$ & $OA$     & $RF$ & J$\&$F  & J${}_{s}$ & J${}_{u}$ & F${}_{s}$ & F${}_{u}$ \\ \midrule
\checkmark   &        &    & 82.7 & 82.1    & 77.4      & 86.3    & 84.9      \\
\checkmark    & \checkmark($W$)   &    & 82.9 & 82.3    & 77.7      & 86.5    & 85.2      \\
\checkmark    & \checkmark($W+S$) &    & 83.6 & 82.4    & 78.7      & 86.8    & 86.5      \\
\checkmark    & \checkmark($W+S$) & \checkmark  & \textbf{83.9} & \textbf{82.6}    & \textbf{79.1}      & \textbf{86.9}    & \textbf{87.1}      \\ \bottomrule
\end{tabular}
}
\label{tab:component_effectiveness}
\end{table}
\textbf{\textit{How does modulator affect the embedding?}} To answer this question, we visualize the transformation process of embedding in $P2C$ over different layers with principal components analysis in \fig{\ref{fig:pca_embedding}}.
We can observe that $P2C$ modulates the embedding by progressively separating the foreground and background features in embedding space.

\paragraph{Modulation with reliable object proxy augmentation.}
To study how the reliable object proxy augmentation influences VOS results, we conduct a set of ablations on YV19-Val split with different proxy construction settings, with $P2C$ as a baseline here. The results are shown in \tab{\ref{tab:component_effectiveness}}.

\textbf{\textit{Does object proxy augmentation make contribution?}} Yes. \tab{\ref{tab:component_effectiveness}} shows that no matter uncertainty patches are filtered or not, using object proxy augmentation always performs better, for both calculating modulation weight and correlation. The reason is that direct propagation may introduce uncertainty while correction from the first frame may be limited due to incomplete semantic guidance. Thus, apart from these two guidance, augmenting object proxy representation especially with reliability helps to complete the semantic concept of an object.

\textbf{\textit{Is reliability-based filtering method beneficial?}} Yes. Comparing $P2C+OA(W+S)+RF$ and $P2C+OA(W+S)$, we can notice that object proxy augmentation with reliability filter outperforms, which indicates that such consideration of reliability during proxy augmentation can further complement with propagation-correction modulators.

\section{Conclusion}
\label{sec:Conclusion}
We present a new modulation-based model that can effectively suppress error propagation in semi-supervised VOS.
The key is to disentangle the correction from the frame-by-frame mask propagation, which provides reliable and comprehensive object proxies as modulation weights and assembles modulators carefully to further consolidate the target embedding.
The object proxy is augmented by supplementing new reliable feature patches from the reliability filter in each iteration, evolving comprehensively. The target embedding encoding the current frame and correlations with references is also consolidated due to the supplemented reliable patches.
The assembly of modulators is critical, and our experiments demonstrate that the cascaded propagation-correction scheme performs the best. 
The main reason is that correction modulation contains global reliable information that could correct errors, and its impact should not be overridden by other modulation. 

We also introduce a reliability filter to facilitate the modulation by assessing prediction quality and selecting reliable feature patches. 
The experiments show impressive gain from the reliable propagation-correction modulation for VOS.

%\bibliographystyle{ACM-Reference-Format}
%\bibliography{sample-base}
\section{Acknowledgement}
\label{sec:Acknowledgement}
The authors would like to thank Xiang Li, Zhaoyang Jia, and Linfeng Qi for meaningful discussion. The authors would also like to thank Rex Cheng for sharing his insightful viewpoints about VOS.  
%This work was supported by the
\bibliography{aaai22}
\clearpage

\setcounter{table}{0}  
\setcounter{figure}{0}  
\renewcommand{\thetable}{\Alph{table}}
\renewcommand{\thefigure}{\Alph{figure}}

\section{Appendix}\label{appendix}
\begin{figure*}[t]
	\centering
	\includegraphics[width=\textwidth]{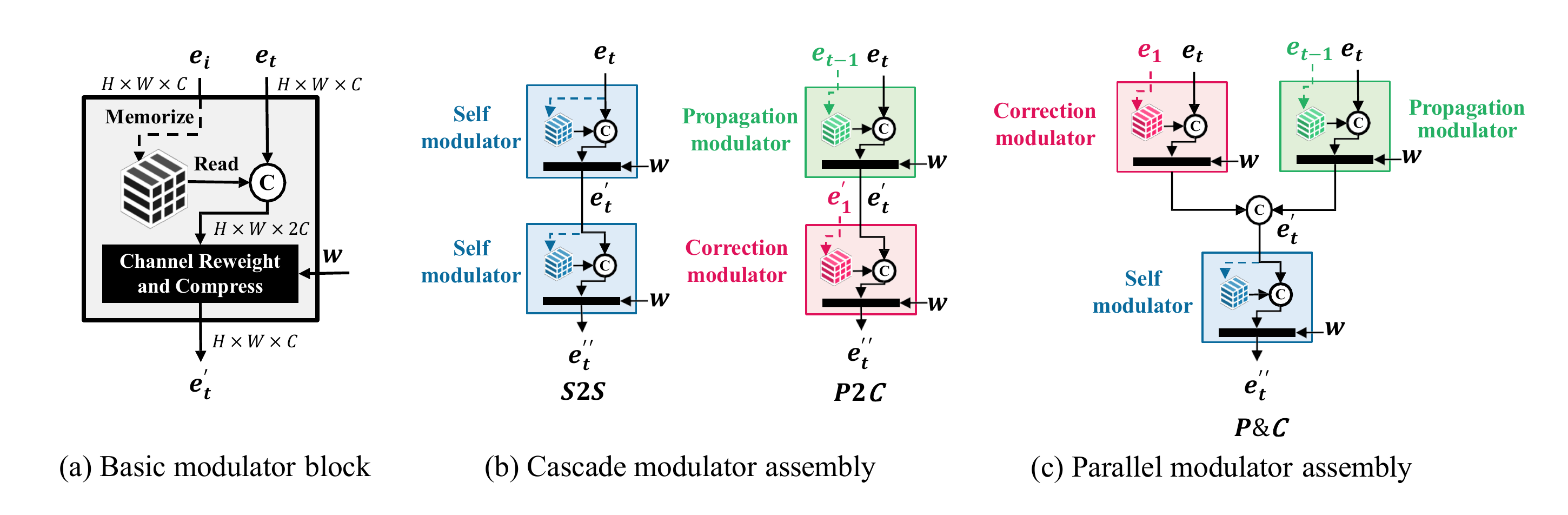}
\caption{(a) Basic modulator block. (b) Cascade modulator assembly scheme. $S2S$ is an assembly with two cascaded self-modulator. $P2C$ stands for propagation-correction cascade scheme. (b) Parallel modulator assembly scheme. We show $P\&C$ here for illustration.} 
	\label{fig:mem_modu_all}
\end{figure*}

In the appendix, we first demonstrate the detailed network structure of Modulator Block. Then, we provide inference speed analysis on DAVIS16 and ablation study for reliability measures. What's more, we provide the qualitative result of a challenging case from the Test-dev split of DAVIS17.

\subsection{Detailed Modulator Block}
\begin{figure*}[ht]
	\centering
	\includegraphics[width=0.8\textwidth]{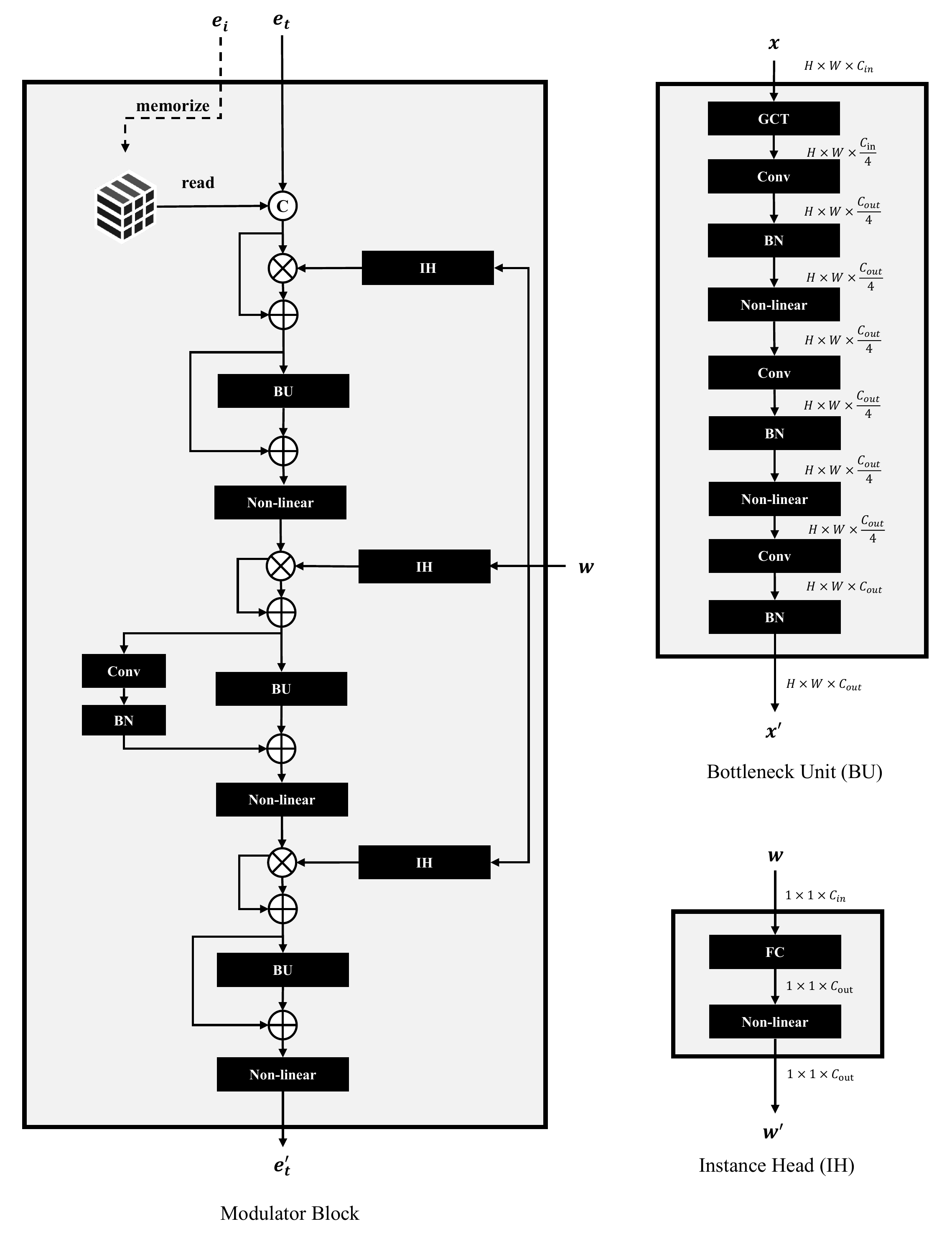}
	\caption{Detailed network structure of a basic modulator block.}
	\label{fig:MemoryModulatorDetail}
\end{figure*}

As shown in Fig.\ref{fig:mem_modu_all}, a modulator block is the basic module to construct various modulator assembly, such as cascade and parallel ones in Fig.\ref{fig:mem_modu_all}(b) and (c).
Fig.\ref{fig:MemoryModulatorDetail} demonstrates the detailed network structure of a basic modulator block. 
A modulator block maintains a memory buffer, which stores memory embedding feature $\mathbf{e}_{i}$ at timestamp $i$ for afterward usage. During each forward operation, the memory modulator block inputs memory embedding features $\mathbf{e}_{t}$ from the current frame, reads out the buffered memory embedding (i.e., $\mathbf{e}_1$ for correction modulator or $\mathbf{e}_{t-1}$ for propagation modulator) and concatenates them together.
Then, several instance heads are introduced at several intermediate layer to re-weight the embedding feature channel-wisely. The instance head includes a fully connected (FC) layer and a non-linear activation function to construct a gate for the embedding feature to be re-weighted. Notably, the spatial resolution of the embedding feature is maintained in the memory modulator block for object detail preservation. What's more, the channel dimension of the input and output feature embedding are also the same. Such design enforces the network to transform and compress the mixture of the feature embedding from current frame and memory bank into a more compact representation. We use Group Normalization (Group Norm) \cite{wu2018group} and Gated Channel Transformation (GCT) \cite{yang2020gated} in the bottleneck unit for stable training.

\subsection{Inference Time Analysis on DAVIS16}
As previous studies \cite{yang2020collaborative,oh2019video,xie2021efficient,oh2019video}, we first compare the inference time of our model with previous state-of-the-art models on DAVIS16 \cite{perazzi2016benchmark}. Then we make an inference time analysis of our reliable proxy augmentation to evaluate whether this component is efficient or not. Our inference protocol mainly follows \cite{yang2020collaborative}, uses one Tesla V100 GPU and set batch size as one. 

\paragraph{Comparison with state-of-the-art models.}
The inference time comparison of our whole model and previous state-of-the-art models is shown in Table \ref{tab:speed_comparison}. Compared to the previous state-of-the-art model CFBI \cite{yang2020collaborative}, whose setting is similar to ours and achieves a good balance of both speed and accuracy, our proposed model not only achieves much better J\&F (90.6\% vs. 89.4\%) but also maintains a faster inference speed (0.172s vs.0.18s). Under the full resolution inference setting, we can further improve the performance from 90.6\% to 91.5\% with little extra time cost (~0.09s). 
\begin{table}[t!]
    \centering
	\caption{Time cost ablation study for reliable proxy augmentation ($RPA$). Here, $P2C$ stands for our propagation-correction modulator scheme. We report the average inference time under two inference resolution settings ($480p$ and Full-resolution $FR$) on DAVIS16 \cite{perazzi2016benchmark}. We first calculate the time $t_{P2C}$ of simply using $P2C$ and the time $t_{P2C+RPA}$ of both using $P2C$ and $RPA$. Then, we calculate the increased time $\Delta{t}$ as $t_{RPA}$ by subtracting $t_{P2C}$ from $t_{P2C+RPA}$.}
	\setlength{\tabcolsep}{6mm}
		\resizebox{0.35\textwidth}{!}
	{
    \begin{tabular}{@{}l|ll@{}}
    \toprule
     Time (s)                            & $480p$   & $FR$ \\ \midrule
    $t_{P2C}$                          & 0.1705 & 0.2625          \\
    $t_{P2C+RPA}$                      & 0.1722 & 0.2628          \\\midrule
    $t_{RPA} (\Delta{t})$                     & 0.0017    & 0.0003        \\
      \bottomrule
    \end{tabular}
    }\label{tab:RPA_time}
\end{table}
\begin{table}[h]
\caption{Ablation study of reliability measures ($M_{r}$) on YouTube-VOS19. $logit$ denotes directly using value of logit map to indicate uncertainty while $SE$ denotes using Shanoon entropy. }

    \centering

    \setlength{\tabcolsep}{3mm}
    \resizebox{0.48\textwidth}{!}
    {
    \begin{tabular}{l|lllll}
    \toprule
       $M_{r}$& J$\&$F  & J${}_{s}$ & J${}_{u}$ & F${}_{s}$ & F${}_{u}$  \\ \midrule
    - & 82.69 & 82.17 & 77.40 & 86.42 & 84.76 \\
     $logit$ & 82.75 & 82.00 &77.50 &86.30 & 85.18  \\
     $SE$  &83.92 & 82.68 & 79.06 & 86.85 & 87.10 \\\bottomrule
    \end{tabular}\label{tab:uncertainty_measure}
     }
     
\end{table}
\begin{table*}[t]
	\centering
	\caption{Quantitative comparison on DAVIS16 \cite{perazzi2016benchmark}. \textbf{$Y$} denotes additionally using YouTube-VOS for training. Superscript $FR$ denotes full-resolution testing. Otherwise, methods are all tested on $480p$. $Ft$ and $S$ separately denote fine-tuning at test time and using simulated data in the training process. We mainly borrow the table from \cite{yang2020collaborative} for inference speed comparison. }
	\setlength{\tabcolsep}{5mm}
		\resizebox{\textwidth}{!}
	{
    \begin{tabular}{@{}lllllll@{}}
    
    \toprule
    Methods    & $Ft$ & $S$ & J\&F & J    & F    & t/s   \\ \midrule
    OSMN \cite{Yang2018osmn}      &   &   &-      & 74.0   &  -    & 0.14  \\
    PML\cite{chen2018blazingly}    &   &   & 77.4 & 75.5 & 79.3 & 0.28  \\
    VideoMatch \cite{hu2018videomatch}&   &   & 80.9 & 81   & 80.8 & 0.32  \\
    FEELVOS(\textbf{Y})  \cite{voigtlaender2019feelvos}&   &   & 81.7 & 81.1 & 82.2 & 0.45  \\
    RGMP \cite{oh2018fast}      &   & \checkmark & 81.8 & 81.5 & 82.0   & 0.14  \\
    A-GAME(\textbf{Y}) \cite{johnander2019generative} &   &   & 82.1 & 82.2 & 82.0   & \textbf{0.07}  \\
    OnAVOS \cite{voigtlaender2017online}    & \checkmark &   & 85.0   & 85.7 & 84.2 & 13    \\
    PReMVOS \cite{luiten2018premvos}    & \checkmark &   & 86.8 & 84.9 & 88.6 & 32.8  \\
    STMVOS \cite{oh2019video}    &   & \checkmark & 86.5 & 84.8 & 88.1 & 0.16  \\
    RMNet(\textbf{Y})  \cite{xie2021efficient}  &   & \checkmark & 88.8 & 88.9 & 88.7 & 0.084 \\
    STMVOS(\textbf{Y}) \cite{oh2019video}  &   & \checkmark & 89.3 & 88.7 & 89.9 & 0.16  \\
    CFBI (\textbf{Y}) \cite{yang2020collaborative}  &   &   & 89.4 & 88.3 & 90.5 & 0.18  \\ \midrule
    \textbf{Ours}(\textbf{Y})       &   &   & 90.6 & 87.1 & 94.0   & 0.172 \\
    \textbf{Ours${}^{FR}$}(\textbf{Y})   &   &   & \textbf{91.5} & \textbf{88.3} & \textbf{94.7} & 0.263 \\ \bottomrule
    \end{tabular}
    }
\label{tab:speed_comparison}
\end{table*} 

\paragraph{Ablation for reliable proxy augmentation}
To validate the efficiency of our reliable proxy augmentation, we further make an inference time ablation study. From Table \ref{tab:RPA_time}, we can notice that the additional time cost by incorporating the reliable proxy augmentation is really slight (about 1\% relative increase under $480p$ resolution inference and 0.1\% relative increase under Full-resolution inference), which further proves the efficiency and effectiveness of this algorithm.

\subsection{Ablation Study for Reliability Measures}

Shannon Entropy ($SE$) is used as a measure of prediction reliability and incorporated in the reliable proxy consolidation. Table \ref{tab:uncertainty_measure} shows the ablation study of different measures of prediction reliability. Apart from using Shannon Entropy, we also tried using the logit map $\mathbf{l}_t$ of each object to directly indicate reliability ($logit$), which brings minor gain.

\begin{figure*}[t]
	\centering
	\includegraphics[width=\textwidth]{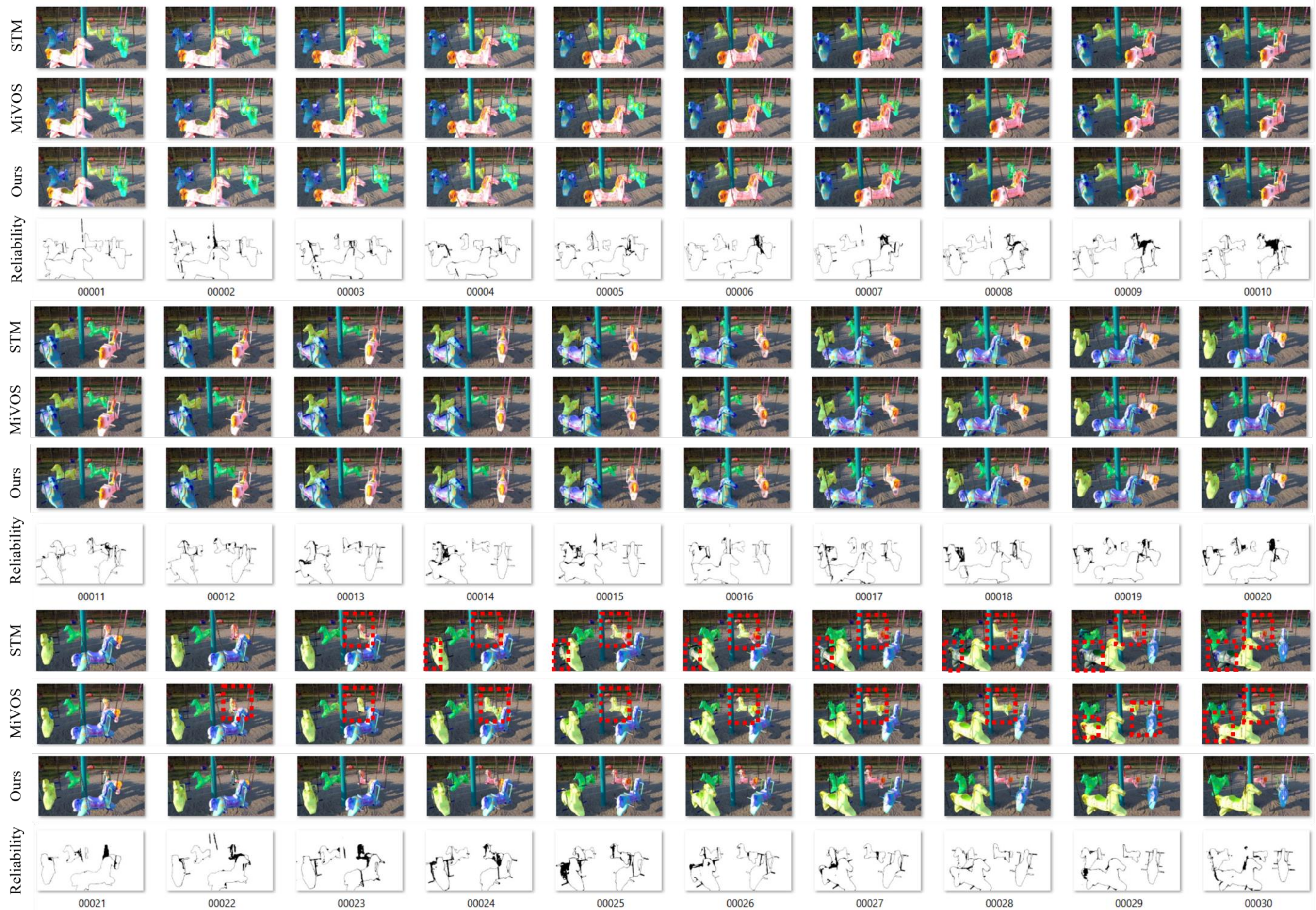} 
	\caption{Qualitative comparison between our model and previous state-of-the-art models on a very challenging case (\textit{carousel}) on DAVIS17 \cite{pont20172017} Test-dev split. }
	\label{fig:qualitative_uncertainty}
\end{figure*}

\subsection{Qualitative Result on DAVIS17 Test-dev}
In Fig.\ref{fig:qualitative_uncertainty}, we show a qualitative comparison between our model and previous state-of-the-art models on a very challenging case (\textit{carousel}) on DAVIS17 \cite{pont20172017} Test-dev split. Considering the large appearance transition  from the reference to targets and interference from similar objects, our model achieves much better results compared to STM \cite{oh2019video} and MiVOS \cite{cheng2021mivos} (red rectangles bound error regions). The binary reliability maps indicate reliable (while) and uncertain (black) patches. From this figure, we can observe that the error regions in a frame of STM and MiVOS tend to propagate into larger ones in the following frames. However, for our model, even if some tiny error regions occur, it can be quickly suppressed in the following frames with the help of reliable guidance.

\end{document}